# On Quantified Linguistic Approximation

**Ryszard Kowalczyk**
CSIRO Mathematical and Information Sciences
723 Swanston Street, Carlton 3053, Australia
ryszard.kowalczyk@cmis.csiro.au

## Abstract

Most fuzzy systems including fuzzy decision support and fuzzy control systems provide outputs in the form of fuzzy sets that represent the inferred conclusions. Linguistic interpretation of such outputs often involves the use of linguistic approximation that assigns a linguistic label to a fuzzy set based on the predefined primary terms, linguistic modifiers and linguistic connectives. More generally, linguistic approximation can be formalized in the terms of the re-translation rules that correspond to the translation rules in explicitation (e.g. simple, modifier, composite, quantification and qualification rules) in computing with words [Zadeh 1996]. However most existing methods of linguistic approximation use the simple, modifier and composite re-translation rules only. Although these methods can provide a sufficient approximation of simple fuzzy sets the approximation of more complex ones that are typical in many practical applications of fuzzy systems may be less satisfactory. Therefore the question arises why not use in linguistic approximation also other re-translation rules corresponding to the translation rules in explicitation to advantage. In particular linguistic quantification may be desirable in situations where the conclusions interpreted as quantified linguistic propositions can be more informative and natural. This paper presents some aspects of linguistic approximation in the context of the re-translation rules and proposes an approach to linguistic approximation with the use of quantification rules, i.e. quantified linguistic approximation. Two methods of the quantified linguistic approximation are considered with the use of linguistic quantifiers based on the concepts of the non-fuzzy and fuzzy cardinalities of fuzzy sets. A number of examples are provided to illustrate the proposed approach.

## 1 INTRODUCTION

The extensive development of fuzzy set theory and fuzzy logic (e.g. [Zadeh 1973, Zadeh 1978, Zadeh 1982, Zadeh 1987]) has led to a general methodology proposed by Zadeh [Zadeh 1996], computing with words. In computing with words the objects of computing are words rather than numbers, with words playing the role of labels of information granules. This methodology covers a wide range of applications of fuzzy systems including fuzzy decision support, decision making, optimization and fuzzy control (e.g. [Zadeh 1973, Zadeh 1996]). The foundations of computing with words lay in well known concepts in fuzzy logic such as a linguistic variable, proposition and granulation, fuzzy inference, fuzzy restrictions (constraints) and fuzzy constraint propagation (e.g. [Zadeh 1973, Zadeh 1975, Zadeh 1978]). In general computing with words involves three main steps [Zadeh 1994, Zadeh 1996]. The first step is explicitation of propositions expressed in a natural language, i.e. representation of the linguistic propositions in their canonical forms. It involves translation of linguistic propositions into the corresponding possibility distributions performed according to the translation rules such as the simple, modifier, composition, qualification and quantification rules [Zadeh 1973, Zadeh 1978, Zadeh 1996]. The second step involves reasoning about the propositions (fuzzy restriction propagation) with the use of the rules of inference in fuzzy logic. The output of this step is a conclusion in the form of a possibility distribution of a fuzzy set. The third step is re-translation of the induced conclusion into a proposition expressed in a natural language, i.e. linguistic proposition. This step involves the use of linguistic approximation that assigns a linguistic label to a fuzzy set.

Many methods of linguistic approximation have been developed and used in both fuzzy decision making [Bonissone 1982, Eshragh and Mamdani 1981, Kowalczyk 1998] and fuzzy control [Novak 1995, Dvorak 1997]. These methods are usually based on combination of predefined primary terms (e.g. small, medium, large), linguistic modifiers or hedges (e.g. not, much, very, more or less) and their connectives (e.g. and, or) that form a linguistic label assigned to a given fuzzy set. For example Bonissone [Bonissone 1982] has developed a linguistic approximation method based on feature extraction and pattern recognition techniques and used it in some problems of decision analysis and natural language processing. A more general approach to linguistic approximation has been proposed in [Eshragh and Mamdani 1981] that uses a combination of segments of the membership function with well defined characteristics. The segments are labeled with the use of linguistic modifiers of the generated primitive terms and the final approximation is a combination of these labels. This technique has been demonstrated for a decision making application [Eshragh and Mamdani 1981]. Similar principles have been used in linguistic approximation presented in [Dvorak 1997] that considers only linguistic terms entering the inference mechanism of a linguistic fuzzy control system [Dvorak 1997, Novak 1995]. A linguistic approximation method based on the use of the principles of evolutionary computation where primary terms, modifiers and connectives are treated as elements of a genetic program has been proposed in [Kowalczyk 1998].

Linguistic approximation can be considered as a complementary task to explicitation and formalized in the terms of the re-translation rules that correspond to the translation rules in explicitation. However it should be noted that most existing methods of linguistic approximation are based on the simple, modifier and composite re-translation rules only. Although these methods can provide a sufficient approximation of simple fuzzy sets the approximation of more complex ones that are typical in many practical applications of fuzzy systems may be less satisfactory. Therefore the question arises why not use in linguistic approximation also other re-translation rules corresponding to the translation rules in explicitation to advantage. In particular linguistic quantification may be desirable in situations where the conclusions interpreted as quantified linguistic propositions can be more informative and natural. This paper presents some aspects of linguistic approximation in the context of the re-translation rules and proposes an approach to linguistic approximation with the use of quantification rules, i.e. quantified linguistic approximation. The principles of re-translation rules in linguistic approximation are presented in section 2. Section 3 proposes two methods of quantified linguistic approximation with the use of linguistic quantifiers based on the concepts of the non-fuzzy and fuzzy cardinalities of fuzzy sets. A number of examples are provided to illustrate the proposed approach. The concluding remarks are presented in section 4.

## 2 RE-TRANSLATION RULES IN LINGUISTIC APPROXIMATION

The fundamental concept used in fuzzy systems and more generally in computing with words is a linguistic proposition [Zadeh 1973, Zadeh 1978, Zadeh 1982, Zadeh 1987, Zadeh 1996]. A simple linguistic proposition takes the form "$X$ is $A$" where $X$ is a variable over the universe of discourse $U$ and $A$ is a linguistic value corresponding to a fuzzy subset of $U$ defined by a membership function $\mu_A$. The variable $X$ has an associated possibility distribution. It is described by a possibility distribution function $\pi_X: U \to [0,1]$ that assigns a degree of possibility to every value of $X$. Translation of linguistic propositions into the corresponding possibility distributions (i.e. explicitation) can be performed according to well known translations rules in fuzzy set theory [Zadeh 1973, Zadeh 1978, Zadeh 1982, Zadeh 1987, Zadeh 1996]. For example in a simple proposition the possibility distribution function of $X$ is equal to the membership function of $A$, i.e.

$$X \text{ is } A \to \pi_X = \mu_A$$

Translation of more complex propositions (e.g. modified, composite, qualified and quantified propositions) involves the use of translation rules such as modifier rules, composition rules, qualification rules and quantification rules [Zadeh 1973, Zadeh 1978, Zadeh 1982, Zadeh 1987, Zadeh 1996]. Examples of the simple, modified and composite linguistic propositions, and their corresponding possibility distributions are presented in figure 1.

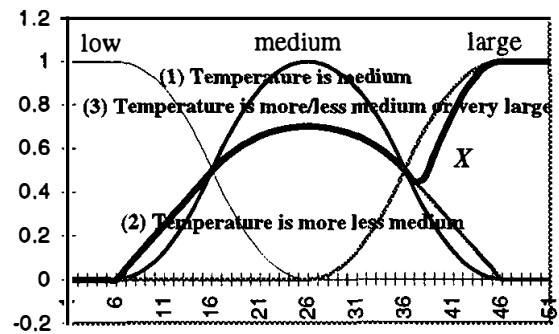

Figure 1: Examples of (1) simple, (2) modified and (3) composite linguistic propositions

A complementary task to explicitation of linguistic propositions in computing with words is re-translation of induced conclusions in the form of possibility distributions into propositions expressed in a natural language, i.e. linguistic propositions. It involves the use of linguistic approximation that assigns a linguistic label to a given fuzzy set.

The problem of linguistic approximation can be defined as mapping from a set $S$ of fuzzy subsets in a universe of discourse $U$, into a set of labels $L$, which are generated according to a grammar $G$ and a vocabulary $V$ [Eshragh and Mamdani 1981]. Typically a solution of linguistic approximation is a linguistic description (label) LA composed of linguistic primary terms $A$, linguistic modifiers $m$ and linguistic connectives $c$ such that it is most suitable (meaningful) to describe a given fuzzy set (a possibility distribution of a linguistic variable). For example a given possibility distribution of a fuzzy set $X$ in figure 1 describing temperature may be linguistically approximated to *"Temperature is more/less medium or very large"*, i.e. LA(X) ≡ X is ($m_1A_1$ c $m_2A_2$) where X ≡ Temperature, $m_1$ ≡ more/less, $A_1$ ≡ medium, $m_2$ ≡ very, $A_2$ ≡ large and c ≡ or. It should be noted that the results of linguistic approximation are not unique and the quality of the provided solutions depends on the error of the approximation expressed typically as the degree of equality of fuzzy quantities [Hirota and Pedrycz 1991, Zwick 1998], i.e. the original fuzzy set and a fuzzy set corresponding to its linguistic approximation [Kowalczyk 1998].

The common characteristic of the existing linguistic approximation methods is that, although not stated explicitly, they generate labels following the principles similar to the translation rules in explicitation. In the context of linguistic approximation these principles can be summarized as the following re-translation rules:

- Simple linguistic approximation
  Given the possibility distribution of a fuzzy set X, its linguistic approximation LA(X) is a simple linguistic proposition as follows:

$$\pi_X \approx \mu_A \rightarrow LA(X) \equiv X \text{ is } A$$

where $\pi_X$ is a possibility distribution function of X, $\mu_A$ is a membership function of a linguistic term A, and ≈ stands for the equality of fuzzy quantities.

- Modified linguistic approximation
  The modifier rule asserts that re-translation of the possibility distribution function is expressed in the the following form:

$$\pi_X \approx \mu_{mA} \rightarrow LA(X) \equiv X \text{ is } mA$$

where $\mu_{mA}$ is a membership function of the modified linguistic term A induced by the linguistic modifier m. In other words m can be interpreted as an operator that transforms the fuzzy set A into the fuzzy set mA. For example if m ≡ very then $\mu_{veryA}(x) = \mu_A^2(x)$.

- Composite linguistic approximation
  The composite re-translation rules apply to linguistic approximation with composite linguistic propositions which are generated from linguistic terms through the use of binary connectives c such as the conjunction (and) and the disjunction (or) as follows

$$\pi_X \approx \mu_{AcB} \rightarrow LA(X) \equiv X \text{ is } A c B$$

For example if c is the conjunction then the composite re-translation rule states that if the possibility distribution of X is equal to the intersection of A and B, i.e. $\mu_{A \wedge B}(x) = \min(\mu_A(x), \mu_B(x))$ then a linguistic approximation of X can be expressed by a composite proposition *"X is A and B"*. It should be noted that the composite re-translation rule can also be applied to more general cases where A and B are defined on two different universes of discourse. More specifically, let U and V be two universes of discourse, and let A and B be fuzzy subsets of U and V, respectively. Then two propositions *"X is A"* and *"Y is B"* connected by the conjunction can be expressed by a composite proposition *"X is A and Y is B"* where the membership function is $\mu_{A \wedge B}(x, y) = \min(\mu_A(x), \mu_B(y))$.

To illustrate the above re-translation rules let us consider two simple problems of linguistic approximation illustrated in figure 2. The task is to assign linguistic labels to two different fuzzy sets $X_1$ and $X_2$ with the use of the simple, modifier and composite re-translation rules and the following elements:

- a set of primary terms T = {small, medium, large}
- a set of linguistic modifiers M = {not, very, more/less, indeed, above, below}
- a set of connectives C = {and, or}

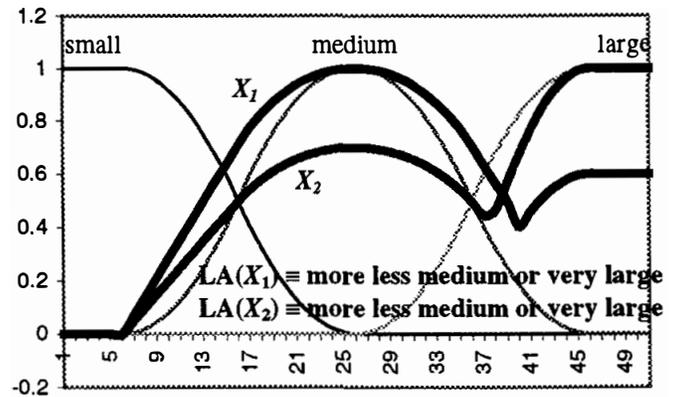

Figure 2: An example of linguistic approximation of two different fuzzy sets

It is easy to observe that both $X_1$ and $X_2$ can be linguistically approximated with the same label, i.e. $LA(X_1)$ = $LA(X_2)$ ≡ *more/less medium or very large*. To distinguish these linguistic approximations one can also provide some additional information expressing the quality of the approximation. There are many measures in fuzzy set theory that may serve the purpose of a measure of the linguistic approximation quality [Hirota and Pedrycz 1991, Yoshikawa and Nishimura 1996, Zwick 1998]. Instances of some potential measures reflecting the difference of fuzzy sets in finite universes that may be applicable in linguistic approximation are listed in table 1. It includes a measure of the height of a fuzzy set, complement of distance such as a normalized Hamming distance, similarity measure and relative sigma count of two fuzzy sets, i.e. the original subnormal fuzzy set (or a segment) and the approximated linguistic label [Kowalczyk 1998]. The last column of the table presents the results of linguistic approximation for examples in figure 2 where the considered measures are applied to each segment of the approximated fuzzy set and the final linguistic approximation.

Table 1: Linguistic approximation with measures [Kowalczyk 1998]

| MEASURE* | | LINGUISTIC APPROXIMATION |
|---|---|---|
| height | $\max \mu_X(x)$ | $LA(X_1) \equiv X_1$ is more less medium (0.70) or very large (1.00) with (1.00) |
| | | $LA(X_2) \equiv X_2$ is more less medium (1.00) or very large (0.60) with (1.00) |
| complement of normalized Hamming distance | $1 - \frac{1}{\#(U)} \sum_{x \in U} \left| \mu_X(x) - \mu_{LA(X)}(x) \right|$ | $LA(X_1) \equiv X_1$ is more less medium (0.67) or very large (0.70) with (0.86) |
| | | $LA(X_2) \equiv X_2$ is more less medium (0.90) or very large (0.47) with (0.91) |
| similarity measure | $1 - \dfrac{\frac{1}{\#(X \cup LA(X))} \sum_{x \in V} \left| \mu_X(x) - \mu_{LA(X)}(x) \right|}{\frac{1}{\#(X)} \sum_{x \in X} \mu_X(x) + \frac{1}{\#(LA(X))} \sum_{x \in LA(X)} \mu_{LA(X)}(x)}$ | $LA(X_1) \equiv X_1$ is more less medium (0.70) or very large (0.71) with (0.89) |
| | | $LA(X_2) \equiv X_2$ is more less medium (0.91) or very large (0.51) with (0.93) |
| relative count | $\dfrac{\frac{1}{\#(X \cap LA(X))} \sum_{x \in V} \min\left( \mu_X(x), \mu_{LA(X)}(x) \right)}{\frac{1}{\#(LA(X))} \sum_{x \in LA(X)} \mu_{LA(X)}(x)}$ | $LA(X_1) \equiv X_1$ is more less medium (0.73) or very large (1.00) with (0.80) |
| | | $LA(X_2) \equiv X_2$ is more less medium (1.00) or very large (0.66) with (0.87) |

\* The measures are calculated for the whole approximated fuzzy set and each of its segments.

In general it can be observed that by providing an additional information about importance of segments and quality of linguistic approximation the approximation measures can enhance the meaning of the generated linguistic labels. In all cases the approximation measures provide the results corresponding with our intuition and expectations. For example the measures in linguistic approximation of $X_1$ indicate that a segment corresponding to the label "more less medium" dominates the segment "very large".

However in some applications it may be desirable to provide more descriptive information about the generated linguistic labels. Following the observation of complementarity of linguistic approximation and explicitation, it seems that other re-translation rules can also be applied in linguistic approximation to advantage. In particular the quantified re-translation rules will be discussed in the next section.

## 3 LINGUISTIC QUANTIFICATION IN LINGUISTIC APPROXIMATION

Quantification is a common means for expressing the scope of propositions. It plays a central role in common-sense knowledge representation and reasoning [Zadeh, 1982, Zadeh 1987]. The classical logic provides two types of quantification, i.e. universal and existential quantification that correspond to the quantifiers *all* and *some*, respectively. Fuzzy logic offers, in addition, a wide variety of fuzzy (linguistic) quantifiers such as few, several, usually, most. These quantifiers are interpreted in fuzzy logic as fuzzy numbers or fuzzy proportions [Zadeh 1982, Zadeh 1987].

The linguistic proposition in the form of $X$ is $A$ (where $A$ can be modified and/or composite) implicitly indicates that it is true for all values of $X$, i.e. *all $X$'s are $A$*. However when only a proportion of values of $X$ satisfies the proposition then the scope specification of this proportion with other quantifiers can be desirable. The quantification rules in explicitation define the translation of quantified linguistic propositions into the canonical forms suitable for further processing such as assessing the truth of a given linguistic proposition. In linguistic approximation quantification can be used to provide the scope of the linguistic labels assigned to the approximated fuzzy set.

In general the quantification rule allows one to consider linguistic quantification in the proposition, i.e. "*QX is A*" where $Q$ is a linguistic quantifier such as many, few, several, all, some, most (e.g. *many X's are large*) as illustrated in figure 3.

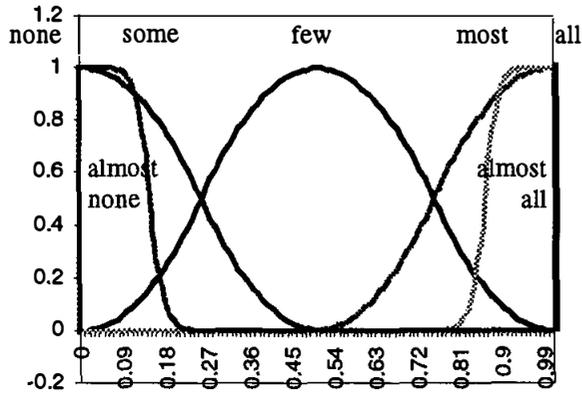

Figure 3: An example of linguistic quantifiers

Referring to the principles of the quantification rules in explicitation [Zadeh 1996], a quantified proposition can be generated in linguistic approximation of a given fuzzy set $X$ with the following quantification re-translation rule:

$$\pi_{Card(A/X)} \approx \mu_Q \rightarrow LA(X) \equiv QX \text{ is } A$$

where $\mu_Q$ is a membership function of a fuzzy set corresponding to the linguistic quantifier $Q$. $Card(A/X)$ denotes the number (or the proportion) of elements of $X$ which are in $A$. In other words it can be considered as cardinality of a fuzzy set corresponding to the intersection of $X$ and $A$. It is well known in fuzzy set theory that this cardinality can be expressed as a non-fuzzy or fuzzy number. More formally, non-fuzzy cardinality of a fuzzy set $F$ is typically defined as a sigma count as follows:

$$\Sigma Count(F) = \sum_{i=1}^{N} \mu_{F(i)}$$

where $\mu_{F(i)}$ is the grade of membership of the $i^{th}$ value of $U$ in the fuzzy set $F$. It should be noted that when $Q$ relates to a proportion (e.g. most) then $\mu_Q$ is a mapping from [0,1] to [0,1], and so-called relative sigma count that expresses the proportion of elements of one fuzzy set $X$ which are in another fuzzy set $A$ can be defined as follows:

$$\Sigma Count(A/X) = \frac{\Sigma Count(A \cap X)}{\Sigma Count(X)}$$

It should be noted that the relative sigma count has been used in some linguistic approximation methods to measure the quality of approximation and to guide the matching process. In the presented approach its use is extended to quantification of the generated label with a linguistic quantifier. Let us consider two problems of linguistic approximation from the previous section (see figure 2) with the additional use of the quantified re-translation rule based on the non-fuzzy cardinality and a set of linguistic quantifiers $Q$ = {none, almost none, some, few, most, almost all, all} as defined in figure 3. To ensure interpretation of the linguistic quantifiers the relative sigma count will be applied to the whole fuzzy set rather than its segments as described in the previous section. Table 2 presents some results of the quantified linguistic approximation for the selected linguistic labels. The linguistic quantifiers are assigned to the labels on the basis of the relative sigma count, i.e. a linguistic quantifier with the highest compatibility degree for a given relative sigma count is selected. In this example it should be noted that *all* elements of the approximated fuzzy sets satisfy the composite label *"more or less medium or very large"* confirming the results of the standard linguistic approximation. In addition the components of such a label can be further described in the quantified linguistic approximation in the terms of linguistic quantification for both fuzzy sets as follows:

$LA(X_1) \equiv$ few $X_1$ are more or less medium;
few $X_1$ are very large

Table 2: Quantified linguistic approximation with non-fuzzy relative sigma count

| Linguistic label $A$ | $\Sigma Count(A/X_1)$ | Linguistic quantifier $Q$ for $X_1$ | $\Sigma Count(A/X_2)$ | Linguistic quantifier $Q$ for $X_2$ |
|---|---|---|---|---|
| small | 0.15 | some | 0.17 | some |
| medium | 0.57 | few | 0.66 | few |
| large | 0.52 | few | 0.34 | few |
| more or less medium | 0.65 | few | 0.83 | most |
| very large | 0.47 | few | 0.28 | few |
| medium or large | 0.97 | almost all | 0.88 | almost all |
| more or less medium or very large | 1 | all | 1 | all |
| | $LA(X_1) \equiv$ few $X_1$'s are more or less medium and few $X_1$'s are very large | | $LA(X_2) \equiv$ most $X_2$'s are more or less medium and few $X_2$'s are very large | |

*LA(X₂)* ≡ *most X₂ are more or less medium;
few X₂ are very large*

It provides more informative description of these fuzzy sets. It should be noted that one can choose to consider more quantifiers to increase granularity of the possible descriptions. In addition a threshold can be included to eliminate linguistic quantifiers that do not satisfy the proportion determined by the relative sigma count above a desired level.

The above interpretation allows one to assign a linguistic quantifier to a generated label during linguistic approximation. It should be noted however that the interpretation of linguistic quantifiers depends on the nature of the propositions considered. For example, if the propositions describe a number of objects then the quantifiers relate to a number (or proportion) of the elements of the approximated fuzzy set that satisfy the proposition (e.g. *many cars are fast, most trucks are heavy*, etc). In general, the use of such quantifiers implies the proposition *QX's are A's*. In addition the linguistic quantifiers can also play a role of assessing the truth of a proposition. It can be referred to as the so-called dispositional quantification with the use of the *usual* and *typical* values of a linguistic proposition [Zadeh 1978]. It is based on a concept of usuality of a proposition meaning that the proposition is usually true (e.g. *usually Temperature is large*). Such quantified propositions can be interpreted as linguistic propositions describing cardinality of a fuzzy set, e.g.

$$usually(X \text{ is } A) \equiv \Sigma Count(A/X) \text{ is most}$$

Therefore quantified linguistic approximation can also be used with propositions in the form of *QX is A*.

Although the non-fuzzy cardinality has commonly been used in explicitation the fuzzy cardinality may be more appropriate in linguistic approximation. In general the fuzzy cardinality of a fuzzy set $A$ is expressed as a fuzzy number as it was proposed in [Zadeh 1982, Zadeh 1987]. More specifically, let $A_\alpha$ be the α-level-set of A, i.e. non-fuzzy set defined by

$$A_\alpha = \{u_i | \mu_A(u_i) \geq \alpha\}, \quad 0 > \alpha \geq 1, u_i \in U, i = 1, \ldots, n$$

where $\mu_i = \mu_A(u_i), i = 1, \ldots, n$ is the grade of membership of $u_i$ in $A$. Then the fuzzy cardinality *FECount(A)* can be represented by intersection of two fuzzy numbers corresponding to the fuzzy cardinalities *FGCount(A)* and *FLCount(A)* describing that at least $n$ elements and at most $n$ elements, respectively are in the fuzzy set $A$ as follows:

$$FECount(A) = FGCount(A) \cap FLCount(A)$$

where

$$FGCount(A) = \sum_\alpha \alpha / Count(A_\alpha)$$

$$FLCount(A) = \sum_\alpha \alpha / Count(\overline{A}_\alpha)$$

where $\Sigma$ stands for the union, *Count* $(A_\alpha)$ denotes the cardinality of the non-fuzzy set $A_\alpha$ and $\overline{A}$ is the complement of the fuzzy set $A$. Similarly the relative fuzzy cardinalities of two fuzzy sets can be defined as follows:

$$FECount(A/X) = FGCount(A/X) \cap FLCount(A/X)$$

where

$$FGCount(A/X) = \sum_\alpha \alpha / \frac{Count(A_\alpha \cap X_\alpha)}{Count(X_\alpha)}$$

$$FLCount(A/X) = \sum_\alpha \alpha / \frac{Count(\overline{A}_\alpha \cap \overline{X}_\alpha)}{Count(\overline{X}_\alpha)}$$

These cardinalities can be used in quantification of linguistic propositions following the principles of the quantified re-translation rules as discussed before. The relative fuzzy cardinalities for the example of linguistic approximation of fuzzy sets $X_1$ and $X_2$ considered before are illustrated in figures 4, 5 and 6. The final quantified linguistic approximation of these sets is based on the assignment of linguistic quantifiers corresponding to the fuzzy cardinality *FECount*. The results confirm that *all*

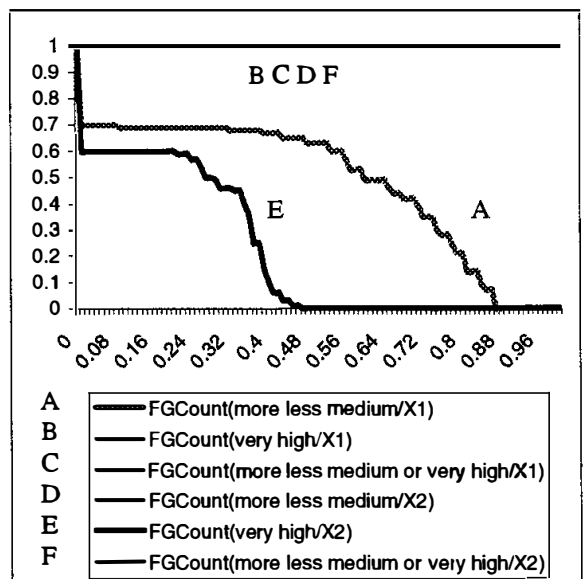

Figure 4: Relative fuzzy cardinality *FGCount*

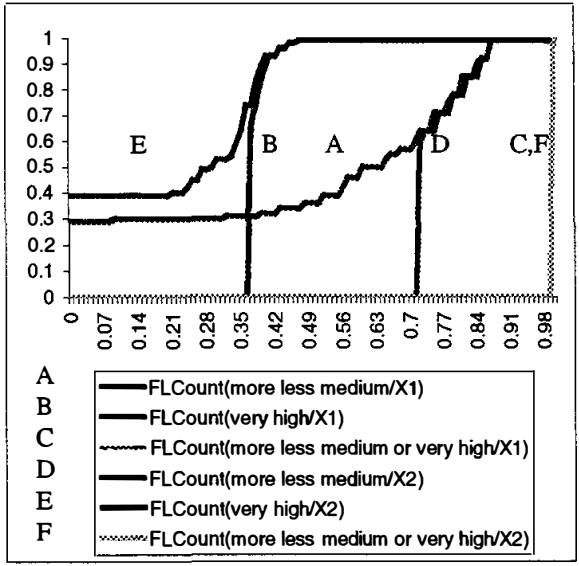

Figure 5: Relative fuzzy cardinality *FLCount*

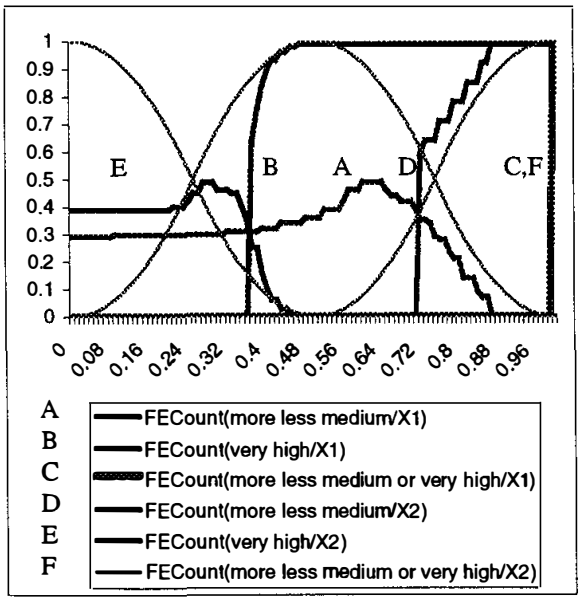

$LA(X_1) \equiv$ few $X_1$'s are more or less medium
and few $X_1$'s are very large
$LA(X_2) \equiv$ most $X_2$'s are more or less medium
and some/few $X_2$'s are very large

Figure 6: Relative fuzzy cardinality *FECount* and quantified linguistic approximation

elements of the approximated fuzzy sets satisfy the composite label "*more or less medium or very large*". In addition the components of this label can be described in the terms of linguistic quantification for both fuzzy sets as follows:

$LA(X_1) \equiv$ few $X_1$ are more or less medium;
few $X_1$ are very large
$LA(X_2) \equiv$ most $X_2$ are more or less medium;
some/few $X_2$ are very large

It should be noted that although the results of quantified linguistic approximation are very similar for both non-fuzzy and fuzzy cardinalities the information provided by the latter seems more meaningful and easier to interpret. In addition, it should also be noted that assignment of a linguistic quantifier to the relative sigma count can be considered as another linguistic approximation problem. However it seems that in this case a simple matching provides sufficient approximation of linguistic quantification.

## 4 CONCLUSIONS

Linguistic approximation has been considered as a complementary task to explicitation in computing with words and formalized in the terms of the re-translation rules that correspond to the translation rules in explicitation. An approach to linguistic approximation with the use of quantification rules, i.e. quantified linguistic approximation has been proposed. Two methods of the quantified linguistic approximation based on the concepts of the non-fuzzy and fuzzy cardinalities of fuzzy sets have been presented and illustrated. Based on the initial results it can be concluded that linguistic quantification can be useful in linguistic approximation to enhance the interpretability of the generated linguistic labels. In particular it seems to be relevant in the problems where information about the scope of the linguistic labels assigned to the approximated fuzzy set is important. It includes commonsense knowledge representation and reasoning in many applications of fuzzy systems for decision support, decision making, optimization and control.

(Eds.) *"Methodology and Tools in Knowledge-Based Systems"*, pp. 200-209.

R. Kowalczyk (1998). On linguistic approximation of subnormal fuzzy sets. *The 17th Annual Conference of the North American Fuzzy Information Processing Society, NAFIPS'98*, August 20-21, Pensacola Beach, Florida.

V. Novak (1995). Linguistically Oriented Fuzzy Logic Controller and Its Design. *Int. J. of Approximate Reasoning*, 12, pp. 263-277.

A. Yoshikawa and T. Nishimura (1996). Relationship between subjective degree of similarity and some similarity indices of fuzzy sets. *Proceedings of IIZUKA'96*, pp.818-821

L.A. Zadeh (1973). Outline of a new approach to the analysis of complex systems and decision processes, *IEEE Trans. Man. Cybernetics*, No.3, pp. 28-44,.

L.A. Zadeh (1978). Fuzzy sets as a basis for a theory of possibility. *Fuzzy Sets and Systems*, vol.1, pp. 3-28,.

L.A. Zadeh (1996). Fuzzy Logic = Computing with Words. *IEEE Transactions on Fuzzy Systems*, vol. 4, no. 2, May, pp. 103-111.

L.A. Zadeh (1982). Test-score semantics for natural languages and meaning representation via PRUF. In B. Rieger (ed.). *Empirical Semantics*. Brokmeyer, Bochum, Germany, pp.281-349,.

L.A. Zadeh (1987). A Computational Theory of Dispositions. *Intern. J. of Intelligent Systems*, 2(1), pp. 39-63,.

R. Zwick, et al. (1988). Measures of Similarity Among Fuzzy Concepts: A Comparative Analysis, *Int. J. Approximate Reasoning*, 1, 221-241.